# AI Deployment and Cyber Governance Failures in Public-Sector Organizations: A Typological Analysis

*Md Salahuddin [1] (✉), James Rooney [2] and Fida Hasan [3]*

[1,3] *School of Professional Studies, University of New South Wales (UNSW), Canberra, Australia*

*m.salahuddin@unsw.edu.au, fida.hasan@unsw.edu.au*

[2]*School of Business, University of New South Wales (UNSW), Australia*

*james.rooney@unsw.edu.au*

**Abstract.**

The intersection of artificial intelligence adoption, cybersecurity governance, and public sector institutional constraints has not been examined as a unified analytical problem in the existing literature. Studies address AI cybersecurity risks generically, public sector governance independently, and framework adequacy separately. Existing studies have not integrated these three streams to explain specifically how AI adoption causes cybersecurity governance failure in government organizations, nor test existing governance instruments against AI-specific public sector failure causes. This paper addresses that gap. It proposes a seven-domain typology identifying ten specific AI-driven cyber governance failure causes grounded in public sector institutional analysis. It presents a three-pathway failure model showing how accountability failure, operational resilience failure, and compliance failure interact and reinforce each other. It delivers a structured coverage matrix testing five major governance frameworks (NIST CSF 2.0, ISO/IEC 27001, COBIT, NIST AI RMF, and ISO/IEC 42001) against the typology, finding that no instrument addresses Shadow AI, speed asymmetry, or governance vacuum at the operational specificity required for public sector application. The paper introduces speed asymmetry as a named structural construct with a specified mechanism. The framework provides the design specification for an AI-enabled cybersecurity maturity model for government organizations.

**Keywords:** cybersecurity governance, artificial intelligence, public sector, Shadow AI, governance failure, maturity model, speed asymmetry, institutional constraints

## 1 Introduction

In 2023, the United Kingdom Cabinet Office issued an emergency notice warning civil servants not to input sensitive official data into generative AI tools [1]. The notice was reactive. Civil servants had already adopted these tools widely across Whitehall for policy drafting and correspondence. Sensitive government data had entered external AI platforms before any governance mechanism existed to prevent it. The failure preceded the policy response by months. Government employees across Australia, the United States, and the EU show the same pattern [2, 3]. Individual adoption moves at the speed of behavior. Governance reform moves at the speed of legislative cycles. The interval between them is ungoverned space where cybersecurity failures accumulate.

Three literature streams address dimensions of this problem without integrating them. AI cybersecurity risk literature documents new attack vectors and operational security challenges [13, 14] without addressing institutional governance failure. Public sector governance literature documents structural weaknesses of government security [8, 9] without linking them to AI adoption specifically. AI governance framework literature evaluates NIST AI RMF and ISO/IEC 42001 [12, 19] without testing them against AI-specific failure causes in the public sector context. No existing instrument integrates all three streams. This paper fills that gap through a structured coverage analysis and proposes an integrated conceptual model to address it.

Four research questions guide this analysis. RQ1: What are the underlying causes of cybersecurity governance failure in public sector organizations arising from the adoption and use of AI? RQ2: How does speed asymmetry between AI adoption velocity and institutional governance reform cycles create structural conditions for cybersecurity governance failure? RQ3: To what extent do existing cybersecurity and AI governance frameworks address AI-specific failure causes in public sector organizations? RQ4: What functional requirements must an AI-enabled cybersecurity maturity model satisfy to address the identified failure causes in public sector contexts?

This paper makes four contributions. The first integrates three research streams that existing literature treats separately: AI adoption behavior, cybersecurity governance structures, and public-sector institutional constraints, analyzed together as a single governance failure problem. The second is original to this paper: speed asymmetry is introduced as a named theoretical construct with a specified mechanism (the structural differential between individual AI adoption speed and institutional governance reform speed), not merely as an informally observed phenomenon present in Taeihagh [34] and Xavier [26]. The third extends prior typological work in cybersecurity and Shadow IT to construct a seven-domain typology that is institutionally attributable, AI-specific, and bounded to the public sector context. The fourth develops an original coverage matrix methodology that tests five major governance frameworks against the typology, producing a demonstration of the governance gap against an external criterion set rather than a self-assessment against each framework's own stated objectives.

Section 2 establishes the conceptual foundations and two-domain governance architecture. Section 3 presents the seven-domain typology identifying ten specific failure causes. Section 4 develops the failure pathway model with cross-domain interaction effects. Section 5 compares existing governance frameworks against typology. Section 6 presents the five-layer conceptual framework. Section 7 discusses theoretical and practical implications. Section 8 concludes.

## 2 Conceptual Foundations and Scope

Cybersecurity governance is the system of policies, roles, and accountability mechanisms through which organizations manage cyber risk [7]. In public sector contexts, it operates within a democratic-legal-ethical accountability frame established by Parliament, courts, the ombudsman, citizens, civil society, and public auditors. This frame sets the mandates, values, and oversight within which all AI governance functions. Pub-lic value, legality, rights, transparency, and accountability are its substantive governance objectives [8].

Within this frame, public sector cybersecurity governance operates under three constraints absent from private sector environments: democratic accountability obligations constrain executive response speed [8]; fixed procurement cycles delay security tool acquisition [21]; and civil service regulations restrict security staffing capacity [21].

AI adoption in government accelerated sharply after 2022. Consumer-accessible generative AI platforms removed prior technical barriers, and government employees began using AI tools for document drafting, data analysis, and correspondence without formal authorization [10, 20, 28, 35, 36]. Governance structures have not kept pace [31].

Shadow AI is employee use of AI tools outside organizational governance awareness, extending the Shadow IT concept documented in peer-reviewed organizational IT research [11, 30]. The consequence is materially different: Shadow IT creates unmanaged IT assets; Shadow AI creates unaudited data flows carrying sensitive government information to foreign-hosted platforms. Empirical evidence of its prevalence is substantial: a 2025 UpGuard survey found over 80 percent of workers use unapproved AI tools, 38 percent have shared confidential work data with AI platforms, and 52 percent have received no AI safety training [23]. Shadow AI is not marginal behavior; governance frameworks have not addressed it. The structural dynamics parallel those of Bring Your Own Device adoption in the early 2010s, documented in Silic and Back [30] and Köffer et al. [11], where employees adopted consumer technology before governance structures existed. The critical difference is that Shadow AI transmits sensitive data to external platforms for processing, rather than using consumer hardware within the organizational perimeter, making the governance consequence qualitatively more severe.

AI-driven governance failure in the public sector arises from two overlapping dimensions. The first is properties intrinsic to AI methods: opacity, automation bias, adversarial susceptibility, and the inferential opacity that prevents audit trails. The second is properties intrinsic to the AI service delivery model: vendor-hosted infrastructure, foreign data processing, third-party model training, and API-based service subscriptions that place sensitive data outside organizational control. The typology in Section 3 maps to both dimensions. AI adoption is not uniform across public sector organizations. Two distinct governance domains operate at different AI maturity levels. The cybersecurity governance domain has integrated AI tools for approximately a decade. Machine learning-based intrusion detection, AI-powered endpoint detection and response, security orchestration and automated response (SOAR), and behavioral anomaly analysis have been operational in government security operations centers since approximately 2015 [13]. This domain has developed partial governance responses and is a forward observation post, encountering AI governance challenges before other parts of the organization face them.

The service delivery and policy governance domain is at a comparatively early stage. AI is now applied to citizen-facing decisions including benefit eligibility assess-ments, case triage, forecasting, policy analysis, and automated correspondence, driven by the availability of commercial generative AI platforms since 2022 [10]. This domain carries high accountability stakes because its AI outputs directly affect citizen rights, public value, and distributive justice. The two domains are not independent. Lessons flow from the mature cybersecurity domain to service delivery: incident discipline, adversarial thinking, and assurance practices. Service delivery AI simultaneously creates new cybersecurity risks (new attack surfaces, sensitive citizen data flows, and supply chain exposures) that feed back into the cybersecurity domain [15, 29].

Public sector organizations operate under four institutional constraints that directly limit technology governance effectiveness [8, 20]. Democratic accountability obligations require transparent, justifiable decisions, slowing executive response cycles. Fixed budget cycles limit investment in new governance capabilities outside annual appropriations. Civil service workforce protections restrict retraining or redeployment in response to new risk categories [21]. Universal service obligations require AI deployments to serve all citizens equitably, regardless of commercial efficiency arguments for narrower scope.

For the purposes of this paper, AI-driven cyber governance failure is defined as: the inability of a public sector organization's governance mechanisms to prevent, detect, attribute, or remediate cybersecurity risks arising from AI deployment or use, including both authorized deployments and unsanctioned Shadow AI adoption. The failure originates in governance mechanisms, not in technical controls. General cybersecurity failures not caused by AI adoption are outside this definition's scope.

## 3 Typology of AI-Driven Cyber Governance Failures

This paper employs a conceptual typological analysis based on a structured review of cybersecurity governance, AI risk management, public-sector digital governance, and Shadow AI literature. Databases consulted included Scopus, Web of Science, IEEE Xplore, and Google Scholar, supplemented by primary framework documentation from NIST, ISO, and ISACA, and advisory publications from the UK Cabinet Office, ASD, and CISA. Sources were selected where they addressed AI-related cybersecurity risk, public-sector governance constraints, or governance framework coverage of AI-specific risks. The typology was constructed by identifying recurring failure mechanisms and retaining only those that were institutionally attributable, AI-specific, and relevant to the public-sector context. The coverage matrix was constructed by mapping each failure cause against explicit provisions in the text of each framework instrument. Search terms combined variants of: AI governance, cybersecurity governance, public sector, Shadow AI, AI risk management, maturity model, and governance framework comparison. Sources were retained where they provided conceptual, empirical, or framework-based evidence relevant to AI-specific cybersecurity governance failure in public-sector organizations. A cause was retained only when traceable to documented evidence and attributable to a mechanism distinct from all others in the typology. The scope is bounded to public sector organizations under democratic accountability constraints.

The analysis identifies ten failure causes organized across seven institutional domains: cross-cutting structural conditions, governance and accountability architecture, procurement and vendor governance, operational and security capacity, workforce behavior and unsanctioned AI, data and sovereignty compliance, and technical opacity and epistemic risk. The seven-domain structure reflects the institutional boundaries within which each cause originates and within which governance interventions must operate. Causes are presented below in domain order, each opening with its name in bold. The analysis presented here is an analytically grounded starting point. Each cause is traceable to documented evidence and attributable to a specific institutional mechanism. Systematic empirical validation through expert interviews, case studies, and a structured Delphi process represents the explicit next stage of the research program. Figure 1 shows the cybersecurity governance as a sub-domain of public service governance with mature AI integration; service delivery and policy as the larger zone where AI integration is emerging. AI enters both at differing maturities (arrow thickness); cybersecurity acts as a forward observation post for the rest of public service, with a recognized augmentation gap (dashed arrow). Ten mediating risks operate across both zones.

**Speed asymmetry** is the structural differential between AI adoption velocity and governance reform speed [10, 26, 34]. The governance-adoption lag is present in prior literature [34]; this paper formalizes it as a named construct with a specified mechanism applicable to the public sector context. Each employee can adopt a new AI tool in minutes. Each governance instrument requires months to years to develop and implement. Speed asymmetry does not produce a specific failure type. It amplifies all other failure causes and widens the interval between risk emergence and governance response. It is the master structural condition within which all other causes operate.

**The governance vacuum** is a structural property of public sector mandate architecture [8, 27]. IT operations, cybersecurity, legal, and procurement each operate within formally defined mandates. AI tool usage by frontline employees falls within none of them. When an AI incident occurs, no unit carries institutional accountability. The vacuum is not negligence; it is the outcome of mandate structures designed before AI adoption became an operational norm

**Absent AI procurement policy.** Standard government procurement processes evaluate cost, vendor stability, and contractual performance but do not evaluate AI model training data provenance, data retention policies, adversarial robustness, or explainability requirements [19, 29]. Security assessments do not occur before deployment. Contractual accountability for AI-specific risks is not established. This absent AI procurement policy is a causal failure at the point of entry, before operational risks materialize.

**Supply chain exposure** is a distinct procurement failure. AI tools are delivered through vendor-managed cloud platforms [10]. A vendor breach constitutes a breach of every government client on that platform. The NIST supply chain risk management frame-work identifies vendor-hosted systems as a primary attack vector in government contexts [29]. Traditional perimeter defense does not address attack vectors inside contracted AI service provider infrastructure. This delivery model constitutes operational outsourcing: AI capability is consumed as a service rather than operated as an asset. Existing supply chain frameworks, including NIST SP 800-161r1 [29] and ISO 27001 supplier controls, were designed for software acquisition, not live model inference accessed through API subscriptions, which creates governance gaps specific to the AI-as-a-service delivery pattern.

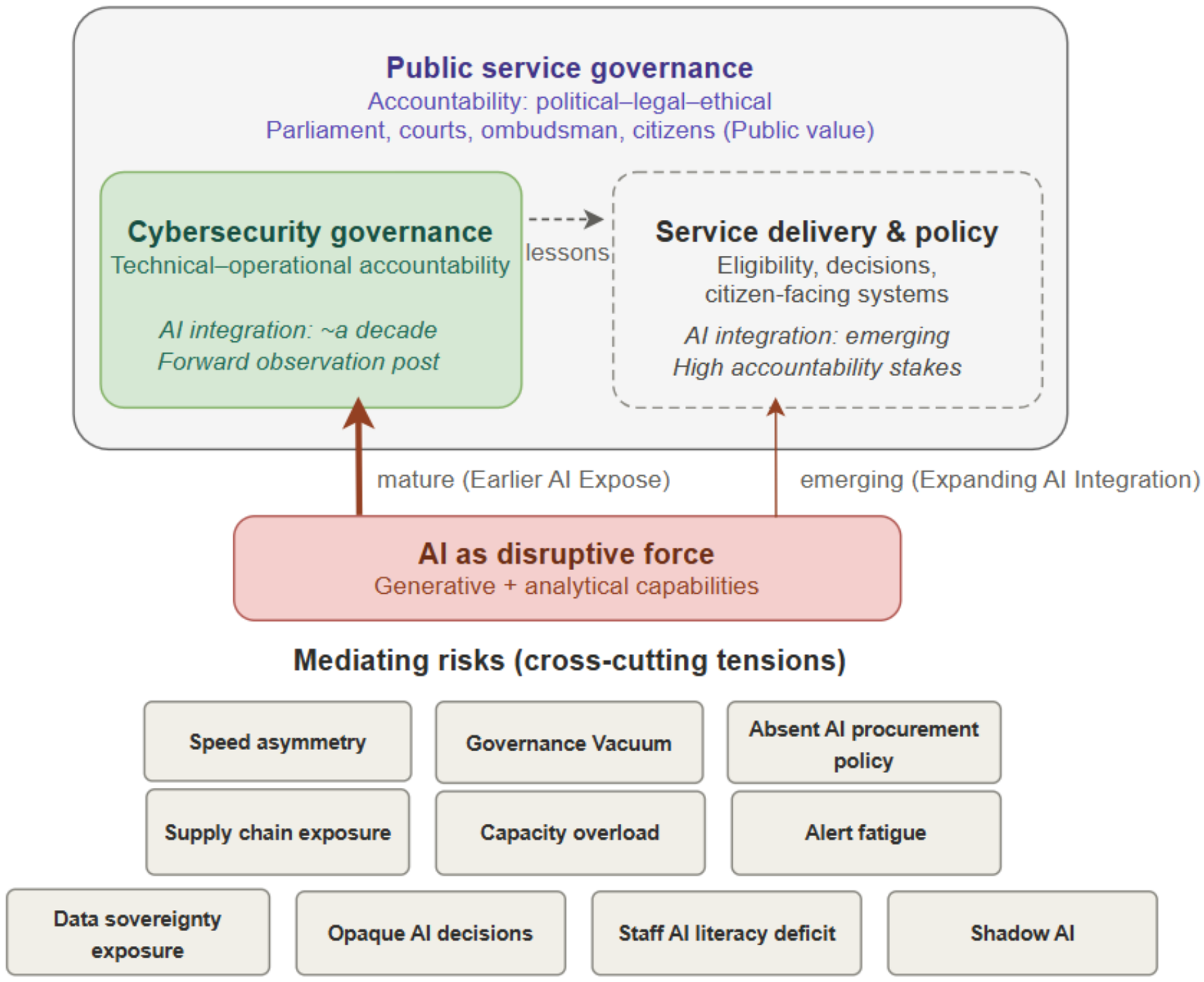

**Fig. 1.** AI as a disruptive force across the two-domain public sector governance architecture, with ten cross-cutting failures causes as mediating tensions.

**Capacity overload** is a structural quantitative mismatch between cybersecurity unit staffing and AI-era threat volumes [24, 25]. Cybersecurity unit staffing is calibrated to pre-AI threat volumes and incident categories. AI deployments increase security alert frequency, expand the organizational attack surface, and generate incident categories that did not previously exist. Units sized for prior-era threat environments cannot manage AI-era incident loads. This is not a competency deficit. It is a scale mismatch between institutional resource allocation and the operational demands of AI-saturated environments.

**Alert fatigue** compounds capacity overload. AI security detection tools generate elevated false positive rates. The SANS SOC Survey 2025 found that 66 percent of SOC teams cannot keep pace with alert volumes [25]. Genuine threats are missed because human triage capacity is exhausted by false positives, not because detection systems fail to flag them [16, 25].

**Shadow AI** is the most operationally prevalent cause in this typology [23, 38]. Government employees use commercial AI tools without security approval, unaware of the data classification obligations that apply to AI platform data retention policies. Shadow AI creates unmapped data flows, unaudited access logs, and attack surfaces outside the organizational security perimeter. Most public sector organizations lack monitoring capabilities to detect these flows.

**Data sovereignty exposure** arises when employees use AI platforms hosted in foreign jurisdictions [1, 10]. National data sovereignty legislation restricts citizen data processing outside approved boundaries. Shadow AI usage routinely breaches these restrictions. The breach is invisible to compliance structures because data movement occurs outside monitored channels. Staff do not recognize the jurisdictional implications of using foreign-hosted AI platforms. The breach is structural in origin, not deliberate non-compliance.

**Technical opacity, epistemic risk, and legitimacy gaps.** Three related failures arise from AI opacity. First, AI systems produce outputs without interpretable reasoning chains [13], which impedes human oversight, forensics, and regulatory accountability. Second, epistemic risk and automation bias mean civil servants accept AI outputs without critical scrutiny [14, 26, 33], producing governance failures in benefit eligibility, risk scoring, and resource allocation even when the AI performs within design parameters. Third, the legitimacy and contestability gap is the citizen-facing consequence: affected citizens cannot understand or challenge AI-influenced decisions through appeals processes not designed for algorithmic reasoning, undermining the democratic accountability architecture [8, 32, 37]. Staff AI literacy deficit amplifies all three; ISC2 2024 found 67 percent of organizations report staffing shortages [24]. Literacy deficit is a systemic infrastructure absence, not a training failure [24].

## 4 AI Governance Failure Pathways

The typology in Section 3 identifies what causes governance failures. This section explains how those causes produce institutional outcomes. The failure pathway model maps the seven typology domains to three governance failure modes: accountability failure, operational resilience failure, and compliance and data governance failure. Speed asymmetry operates across all three pathways, widening the gap between technological capability and governance response. Figure 2 presents the complete framework.

Accountability failure occurs when AI incidents cannot be attributed to a responsible unit, contractual owner, or audit trail. Its primary security consequence is the inability to conduct post-incident forensic attribution: organizations cannot determine the origin, scope, or responsible party for AI-related breaches, preventing any corrective action or legal remedy. Governance vacuum, absent vendor accountability, and Shadow AI combine to produce this outcome structurally [8, 27].

Capacity overload increases Mean Time to Detect (MTTD) and Mean Time to Respond (MTTR), degrading incident response performance [16, 25]. Alert fatigue degrades analyst decision quality. Adversarial AI targets detection windows [14, 22]. Supply chain breaches bypass perimeter controls [15, 29]. Resilience failure manifests as prolonged breach windows and delayed recovery.

Compliance failure produces three specific security and legal consequences. Shadow AI generates data sovereignty breaches outside compliance awareness [1, 5], exposing organizations to regulatory liability under data protection legislation. Opaque AI prevents the decision audit trails that legislation such as GDPR Article 22 requires [13, 14], producing systematic documentation failures. Staff literacy deficits prevent behavioral compliance [23, 24], ensuring that policy commitments are not translated into operational behavior. These outcomes arise from a structural gap between policy intent and operational reality.

The three pathways compound each other. Accountability failure prevents correct activation of resilience and compliance procedures [8, 27]. Resilience failure prevents the systematic monitoring that compliance requires [16, 25]. Compliance failure consumes the management attention that reform requires. These are not independent failure modes; they operate as a mutually reinforcing system in which speed asymmetry amplifies all three loops simultaneously, as governance reform accumulates more slowly than failure causes generate new incidents [10, 26, 34].

## 5 Comparison with Existing Governance Frameworks

This section assesses five frameworks against the typology using three ratings. Full: explicit operational controls or accountability structures for the cause. Partial: relevant general guidance requiring substantial organizational adaptation. None: no relevant provision identified in the framework text.

NIST CSF 2.0 addresses supply chain risk management explicitly but does not define AI-specific controls, address Shadow AI, or provide accountability structures for unsanctioned employee AI usage [4]. ISO/IEC 27001:2022 covers supplier relationships (A.5.19–A.5.22) and data protection (A.5.31) but no control addresses AI tool deployment by non-IT staff [5]. AI tools used through consumer interfaces do not appear on standard asset registers. COBIT 2019 assumes IT decisions flow through formal governance channels, an assumption that does not hold for consumer AI adoption by frontline staff [6].

The NIST AI RMF organizes AI risk through Govern, Map, Measure, and Manage functions [19]. Its Govern function provides the strongest procurement-related coverage of any reviewed framework. Its Explain function addresses algorithmic explainability fully. However, the AI RMF is designed for AI system developers and deployers, not government organizations managing employee-led AI adoption. It does not address

Shadow AI or public sector mandate architecture. ISO/IEC 42001:2023 provides strong explainability and transparency coverage and partially addresses procurement governance [12]. Its scope is AI system providers and operators. A government ministry managing consumer AI adoption by staff is outside its stated application context. Neither framework addresses Shadow AI, governance vacuum, or speed asymmetry. National agency guidance remains equally partial: UK Cabinet Office [1] addresses Shadow AI at policy level without operational controls, ASD guidelines [2] cover deployment security but not employee-led adoption, and CISA [15] addresses adversarial AI without institutional governance failures. Each instrument covers one dimension of the typology, not the full architecture.

Table 1 summarizes the framework coverage assessment across all five instruments and ten failure causes, using ratings of Full, Partial, and None as defined at the start of this section.

**Table 1.** Table captions should be placed above the tables.

| Failure Cause | NIST CSF 2.0 | ISO 27001 | COBIT | NIST AI RMF | ISO 42001 |
|---|---|---|---|---|---|
| Speed asymmetry | None | None | None | Partial | None |
| Governance vacuum | Partial | Partial | Partial | Partial | Partial |
| Absent AI procurement | Partial | Partial | Partial | Full | Partial |
| Capacity overload | Partial | None | None | None | None |
| Alert fatigue | None | None | None | None | None |
| Supply chain exposure | Full | Full | Partial | Partial | Partial |
| Shadow AI deployment | None | None | None | None | None |
| Data sovereignty | Partial | Full | Partial | Partial | Partial |
| Opaque AI decisions | None | None | None | Full | Full |
| Staff AI literacy | Partial | Partial | Partial | Partial | Partial |

Key rating decisions warrant explanation. NIST CSF 2.0 receives Full for supply chain exposure because its Supply Chain Risk Management category (SC.AS-01, SC.AS-02, RS.MA-04) provides explicit third-party risk controls applicable to AI infrastructure. Speed asymmetry receives None from NIST CSF, ISO 27001, and COBIT because none provides any mechanism for the governance-adoption temporal differential; NIST AI RMF receives Partial because its GOVERN function (GV.OC-04) addresses keep pace with AI evolution without specifying how. Governance vacuum receives Partial across all five frameworks because each addresses accountability in general terms (e.g., ISO 27001 A.5.2, COBIT APO01) without defining structures for em-ployee-led unsanctioned AI adoption. Shadow AI receives None from every framework reviewed; no instrument provides detection controls, approved-use registers, or governance requirements for unsanctioned employee AI tool use. These are verified against the text of each framework document. Taken together, Table 1 shows supply chain exposure and data sovereignty as the best-covered causes, while Shadow AI, speed asymmetry, governance vacuum, and alert fatigue remain unaddressed across all five instruments. The reviewed instruments do not integrate accountability, resilience, and compliance governance for AI adoption in a single instrument tailored to public-sector organizations. Section 6 addresses that gap.

## 6 Conceptual Framework

Sections 3, 4, and 5 identify the causes, map the failure pathways, and show why existing frameworks fall short. This section draws those threads into a single conceptual model structured across five layers. Figure 2 presents the complete framework.

The outermost layer is the constitutional accountability frame: Parliament, courts, the ombudsman, citizens, and public auditors set mandates, risk appetite, and oversight for every layer below. Public value, legality, rights, transparency, and accountability are its substantive governance objectives [8]. A governance instrument that does not serve these objectives lacks democratic legitimacy, whatever its technical merits. Spanning both operational domains beneath this frame is the AI Governance and Assurance Bridge, covering six governance lifecycle functions: Govern, Map, Measure, Manage, Assure, and Respond and Redress. Existing frameworks do not cover all six at the specificity required for public sector application, as Table 1 demonstrates.

The cybersecurity governance domain has approximately a decade of AI integration experience and acts as a forward observation post for the organization. The service delivery and policy domain carries emerging, high-accountability-stakes AI integration in citizen-facing decisions: eligibility, triage, casework, and forecasting. These domains exchange knowledge and risks bidirectionally. Lessons and risks flow both ways, and neither domain can be governed in isolation [26, 27].

The ten failure causes identified in Section 3 operate as structural tensions across both domains in Layer 4. Speed asymmetry amplifies all others by ensuring governance responses always lag adoption. Layer 5 contains the foundational preconditions: data governance, infrastructure, security controls, workforce skills, procurement, and legacy systems. These are not governance mechanisms; they are the conditions that determine what governance mechanisms can achieve.

External actors (vendors, cloud providers, regulators, and other agencies) interact with both operational domains across organizational boundaries. A public feedback and accountability loop (complaints, appeals, audits, and parliamentary scrutiny) feeds into the democratic frame at Layer 1. This loop makes the legitimacy and contestability gap institutionally visible. The framework positions an AI-enabled cybersecurity maturity model as the bridging instrument across all five layers.

## Maturity Model Requirements: Answer to RQ4

RQ4 asks what functional requirements an AI-enabled cybersecurity maturity model must satisfy. Following the maturity model design methodology established in the information systems literature [17, 18], Table 2 provides those requirements, derived directly from the typology. Each requirement specifies what the maturity model must assess or govern: from initial governance awareness through to institutionalized governance capacity.

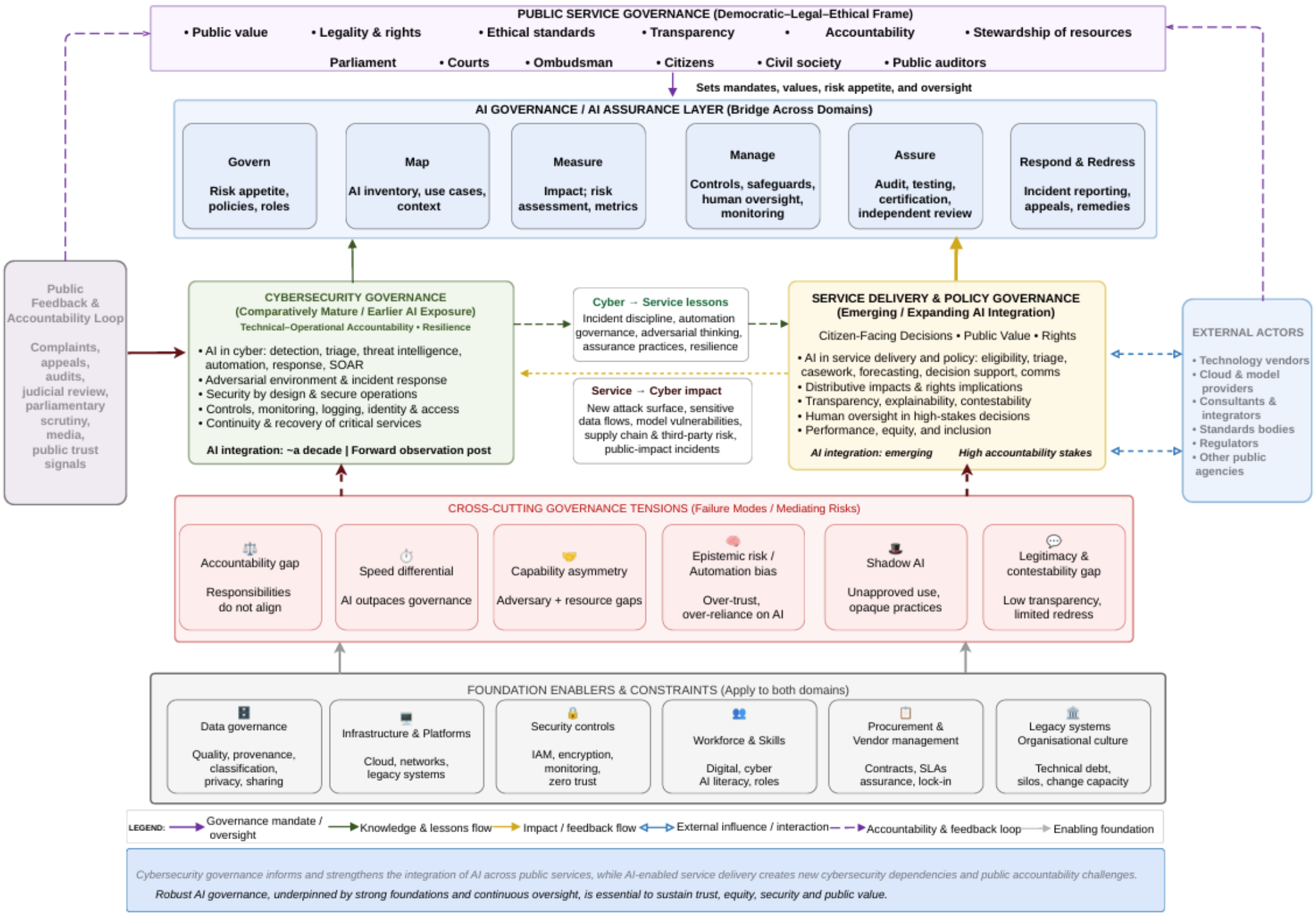

Fig. 2. Five-layer AI governance framework for public-sector organizations, showing the relationship between democratic accountability, AI assurance, cybersecurity governance, public-service delivery, cross-cutting failure causes, and foundation enablers and constraints.

**Table 2. Functional requirements for an AI-enabled cybersecurity maturity model, derived from the seven-domain typology.**

| Failure Cause | AI-Enabled Cybersecurity Maturity Model Requirement |
|---|---|
| Speed asymmetry | Continuous AI-use inventory mechanism and rapid governance review cycle aligned to AI adoption pace |
| Governance vacuum | Explicit AI accountability role assignment and mandate extension covering employee-led AI adoption |
| Absent AI procurement | AI-specific vendor assurance criteria and pre-deployment risk assessment integrated into procurement |
| Supply chain exposure | Third-party AI provider security controls, contractual obligations, and ongoing assurance monitoring |
| Capacity overload | AI-era staffing benchmarks, workload metrics, and scalable incident response capacity standards |
| Alert fatigue | Structured AI-alert triage protocols, false positive management standards, and analyst workload limits |
| Shadow AI | Shadow AI detection capability, approved AI-use policy, and staff reporting mechanisms |
| Data sovereignty | Data classification controls, approved AI-platform register, and jurisdictional boundary enforcement |
| Opaque AI decisions | Explainability standards, AI decision audit trail requirements, and human oversight checkpoints |
| Staff AI literacy | AI literacy assessment framework, training standards, and role-based competency benchmarks |

## 7 Discussion

### Theoretical Contributions

AI cybersecurity risk, public sector governance failure, and framework adequacy have each been treated as separate research problems [8, 13, 19]. This paper treats their intersection as a single analytical object.

Speed asymmetry is formalized as a named construct with a specified mechanism: the differential between individual AI adoption speed and institutional reform speed. This makes it testable, replicable, and applicable beyond the public sector.

Table 1 shows that no reviewed framework addresses Shadow AI, speed asymmetry, or governance vacuum at the operational specificity the public sector context requires [4, 5, 6, 12, 19]. This finding is grounded in direct analysis of each framework's text rather than in self-reported coverage.

### Practical Implications for Public-Sector Organizations

The typology can function as a self-assessment instrument. Speed asymmetry and governance vacuum require mandate redesign. Shadow AI requires approved-use policies, staff education, reporting channels, and technical controls to provide organizational visibility over unauthorized AI tool use. Table 1 enables targeted investment: NIST

CSF provides supply chain coverage; ISO 27001 covers data sovereignty; neither addresses Shadow AI, speed asymmetry, or governance vacuum.

**Implications for Cybersecurity Governance Practice**

Cybersecurity teams must formally extend their mandate to include AI governance functions, not absorb additional tasks on existing capacity. Section 3 shows capacity overload is already occurring at current AI deployment levels [24, 25]. Mandate expansion without resource expansion will accelerate the resilience failure pathway, not resolve it.

**Ethical and Public-Value Considerations**

Citizens interact with AI-enabled government services without choosing to do so. They cannot opt out of an AI-processed benefit decision or an AI-triaged service request. This asymmetry creates governance obligations that extend beyond technical security to encompass fairness, transparency, and accountability as substantive requirements [8, 32]. The service delivery domain carries the highest concentration of these obligations because its AI outputs directly affect citizen rights, benefit access, and distributive justice.

The legitimacy and contestability gap identified in Section 3 is the ethical failure when these obligations are unmet. Citizens who cannot understand or challenge AI-influenced decisions experience a breakdown in the democratic compact between state and citizen. This is a public value failure, not only a compliance risk [8].

**Limitations**

This paper is a conceptual contribution. The typology and pathway model are derived from literature synthesis and analytical judgement, not from systematic empirical data collection such as expert interviews, case studies, or a formal coding process. This defines the agenda for subsequent empirical work rather than disqualifying the analytical contribution: analytically grounded typologies are an established first contribution in governance research against which empirical validation can then be conducted. The typology may not capture causes specific to military, intelligence, or developing-economy government contexts. The coverage matrix assessments represent analytical judgements based on published framework documentation; practitioners may achieve different coverage levels through supplementary guidance and organizational adaptation.

**Future Research Directions**

Future work should validate the typology empirically through case studies and expert interviews. Cross-national studies examining how political structures modulate failure causes would strengthen generalizability [26, 27]. The next stage is to design, operationalize, and field-test an AI-enabled cybersecurity maturity model grounded in this framework.

## 8 Conclusion

This paper argued that AI-driven cybersecurity governance failure in public sector organizations is a distinct institutional problem that cannot be resolved by existing tech-nical risk frameworks or generic AI governance instruments. By analyzing the institutional mechanisms through which AI adoption produces governance failure, the study identified ten specific failure causes across seven domains, showed analytically how these causes interact through three reinforcing failure pathways, and showed through a structured coverage matrix that no existing governance framework addresses Shadow AI, speed asymmetry, or governance vacuum at the operational specificity the public sector context requires. The central construct introduced is speed asymmetry, the structural differential between AI adoption velocity and institutional governance reform speed, which amplifies every other failure cause in the typology. The key implication is that governments cannot resolve AI-driven cybersecurity governance failures through technical controls alone. They must redesign accountability mandates, procurement structures, workforce literacy infrastructure, and data sovereignty controls as first-order institutional governance priorities. Framing AI-driven cybersecurity failure as an institutional governance problem rather than a technical risk problem shifts the analytical frame for public sector cybersecurity governance from technical risk management to institutional governance design and provides the design specification for an AI-enabled cybersecurity maturity model capable of assessing governance readiness, accountability, resilience, and compliance in government organizations.

### Disclosure of AI Assistance

The intellectual contributions in this paper, including the typology, failure pathway model, conceptual framework, and coverage matrix, were developed entirely by the authors. Generative AI tools were used solely to improve grammatical accuracy and sentence structure in the preparation of the manuscript. No AI tool contributed to the research design, data analysis, or conclusions. The author takes full responsibility for the integrity and accuracy of all content.

### Disclosure of Interests

The authors have no competing interests to declare that are relevant to the content of this article.